\documentclass[letterpaper]{article} 
\usepackage{aaai2027}  
\usepackage[hyphens]{url}  
\usepackage{graphicx} 
\usepackage{natbib}  
\usepackage{caption} 
\usepackage{algorithm}
\usepackage{algorithmic}
\usepackage{multirow}
\usepackage{newfloat}
\usepackage{listings}
\DeclareCaptionStyle{ruled}{labelfont=normalfont,labelsep=colon,strut=off} 
\floatstyle{ruled}
\newfloat{listing}{tb}{lst}{}
\floatname{listing}{Listing}

\usepackage{booktabs}

\title{Rethinking Auxiliary Modalities in Multimodal Zero-shot Anomaly Detection: From Semantic Fusion to Conditional Modulation}
\author{
    Peng Wu\textsuperscript{\rm 1},
    Xin Ge\textsuperscript{\rm 1},
    Yujia Sun\textsuperscript{\rm 2}\thanks{Corresponding author},
    Guansong Pang\textsuperscript{\rm 3}
}

\affiliations{
    \textsuperscript{\rm 1}School of Computer Science, Northwestern Polytechnical University, China\\
    \textsuperscript{\rm 2}School of Artificial Intelligence, Xidian University, China\\
    \textsuperscript{\rm 3}School of Computing and Information Systems, Singapore Management University, Singapore\\
    xdwupeng@gmail.com, gx6173@gmail.com, yjsun@stu.xidian.edu.cn, gspang@smu.edu.sg
}

\begin{document}

\maketitle

\begin{abstract}
Recent foundation model-based methods have endowed RGB images with strong zero-shot anomaly detection (ZSAD) through vision-language pretraining. However, RGB observations alone remain limited in perceiving anomalies dominated by geometric deformation, depth variation, or subtle surface changes. Auxiliary modalities can provide complementary structural information, but existing multimodal methods typically fuse them directly into a shared semantic space, which may disturb the text-aligned anomaly semantics established by RGB foundation models and often requires modality-specific architectures. 
To address this issue, we propose a plug-and-play auxiliary-conditioned enhancement framework for zero-shot anomaly detection. Instead of reconstructing a joint multimodal anomaly semantic space, 
our framework preserves the original RGB image-text anomaly matching pathway and uses auxiliary observations as conditional signals for RGB feature refinement, allowing auxiliary modalities to seamlessly enhance existing RGB-based zero-shot anomaly detectors. 
Specifically, a lightweight meta-learning module takes global RGB and auxiliary representations as input and generates sample-adaptive low-rank residual updates to determine how RGB features should be refined. We further construct uncertainty-aware spatial modulation from the initial RGB anomaly response and auxiliary reliability, which determines where local residual updates are strengthened or suppressed. This global-to-local conditional modulation enables selective multimodal enhancement while preserving the original RGB anomaly semantics. Extensive experiments on MVTec 3D-AD and Eyecandies demonstrate that our framework consistently improves multiple popular RGB-based zero-shot anomaly detectors, achieving state-of-the-art performance for multimodal zero-shot anomaly detection.

\end{abstract}

\section{Introduction}
Image anomaly detection~\cite{yang2023memseg, chen2026dyc, li2026iad} aims to identify anomalous regions or samples that deviate from normal patterns, playing a critical role in industrial inspection applications. Since anomalous samples are inherently scarce and difficult to exhaustively collect, conventional methods primarily learn representations of normality from normal data, such as reconstruction patterns or memory prototypes, and detect anomalies as deviations from the learned normal patterns~\cite{wu2026deep}. However, these methods heavily rely on the coverage of normal data and often struggle to generalize to unseen anomaly categories. Recently, vision-language foundation models have introduced a new paradigm for anomaly detection. By leveraging the open-set semantic knowledge acquired through large-scale vision-language pretraining, they identify unseen anomalies via image-text matching, shifting anomaly detection from normality modeling to foundation model-based zero-shot recognition.

\begin{figure*}[t]
\centering
\includegraphics[width=0.85\textwidth]{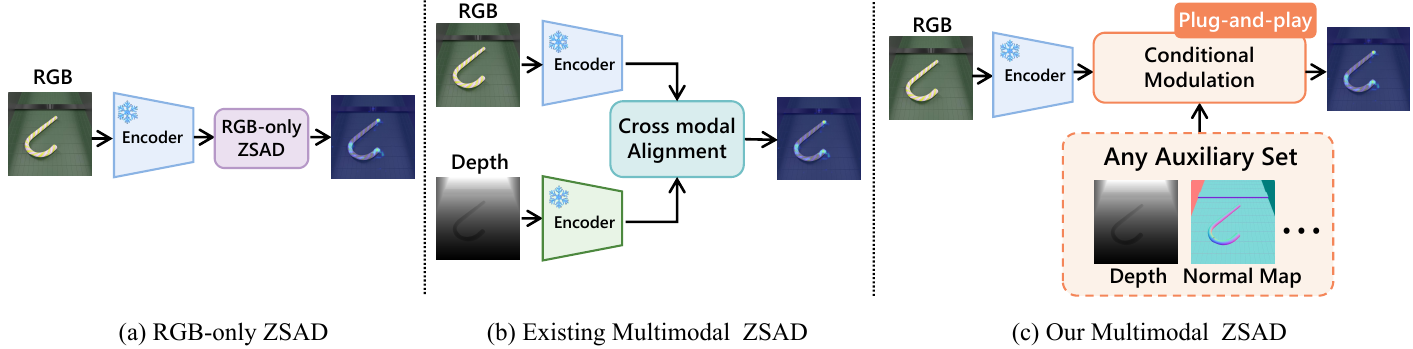}
\caption{Comparison of zero-shot anomaly detection paradigms. 
(a) RGB-only ZSAD relies on RGB image-text matching. 
(b) Existing multimodal ZSAD methods align RGB and auxiliary features in a shared semantic space. 
(c) Our method uses available auxiliary subsets as conditional modulation signals for plug-and-play enhancement.}
\label{fig:paradigm}
\end{figure*}

Although foundation models substantially expand the open-set capability of anomaly detection~\cite{wu2024open}, existing zero-shot methods still rely primarily on RGB image-text alignment~\cite{ma2025aa, yuan2026ad}. Due to the inherent limitations of appearance-based imaging, RGB observations may fail to capture anomaly cues associated with geometric deformation, surface orientation, or depth variation, especially under weak texture contrast or complex illumination. Recent multimodal industrial benchmarks therefore introduce complementary observations, such as depth maps, surface normal maps, and 3D geometry, to provide structural and physical cues beyond RGB, as illustrated in Figure~\ref{fig:paradigm}(a). However, the emergence of foundation models also changes the role of auxiliary modalities. Since RGB foundation models have already learned powerful open-set anomaly semantics, the key challenge is no longer how to jointly construct multimodal anomaly semantics, but how to effectively leverage auxiliary modalities to enhance RGB anomaly perception without disrupting the established vision-language semantic space.

As illustrated in Figure~\ref{fig:paradigm}(b), existing multimodal zero-shot anomaly detection (ZSAD) methods typically extract RGB and auxiliary features separately and integrate them through cross-modal alignment, shared representation learning, or dynamic feature interaction. Despite their different implementations, these methods share a common assumption that RGB and auxiliary modalities contribute equally to anomaly semantic modeling by jointly constructing a unified multimodal representation. However, this assumption becomes questionable in the era of foundation models. RGB foundation models have already established powerful open-set anomaly semantics through vision-language pretraining, whereas auxiliary modalities primarily provide complementary structural and physical cues that are difficult to capture from RGB alone, rather than independent text-aligned anomaly semantics. Treating auxiliary modalities as additional semantic sources may therefore interfere with the well-established RGB vision-language semantic space and introduce unnecessary cross-modal representational bias. We argue that the role of auxiliary modalities should be redefined: instead of jointly constructing anomaly semantics, they should serve as complementary perceptual cues that selectively enhance RGB anomaly perception while preserving the original RGB image-text semantic alignment.

Based on this observation, we propose a plug-and-play auxiliary-conditioned framework for foundation-based multimodal zero-shot anomaly detection, as illustrated in Figure~\ref{fig:paradigm}(c). Rather than treating auxiliary modalities as additional sources of anomaly semantics, our framework preserves the original RGB image-text matching pathway and leverages auxiliary observations solely as conditional signals to refine RGB anomaly perception. This design enables existing RGB-based zero-shot anomaly detectors to exploit complementary structural and physical cues without reconstructing a joint multimodal anomaly semantic space or redesigning their detection pipelines. Specifically, we formulate auxiliary enhancement as a global-to-local modulation process. At the global level, a lightweight meta-learning network takes RGB and auxiliary representations as input and generates sample-adaptive Low-Rank Adaptation (LoRA) parameters for selected Transformer layers, determining how RGB features should be refined. At the local level, uncertainty-aware spatial modulation combines the initial RGB anomaly response and an auxiliary reliability map to determine where the generated residual updates should be strengthened or suppressed. By separating the refinement direction from spatial modulation, the proposed framework avoids indiscriminate full-image fusion and enables lightweight and spatially selective enhancement across different available auxiliary modalities. Moreover, experiments demonstrate that the proposed plug-and-play framework can be seamlessly integrated into multiple popular RGB-based zero-shot anomaly detectors, including AnomalyCLIP \cite{zhou2024anomalyclip}, AA-CLIP \cite{ma2025aa}, and AnomalyVFM \cite{fucka2026anomalyvfm}, thereby consistently improving their performance. Our contributions are summarized as follows:
\begin{itemize}
\item We propose a plug-and-play auxiliary-conditioned enhancement framework for foundation-based multimodal zero-shot anomaly detection. It recasts auxiliary modalities from direct anomaly semantic sources as conditional signals for RGB anomaly perception, thereby avoiding the construction of a joint multimodal anomaly semantic space and enabling seamless integration with existing RGB-based zero-shot anomaly detectors.

\item We propose a global condition-driven dynamic LoRA modulation mechanism. It derives joint conditions from global RGB and auxiliary representations and employs a lightweight meta-learning network to generate sample-adaptive low-rank residual updates, which determine how RGB features are refined for each input.

\item We propose an uncertainty-aware spatial modulation mechanism, which combines RGB prediction uncertainty and auxiliary reliability to adaptively strengthen or suppress local residual updates.

\item Extensive experiments on MVTec 3D-AD and Eyecandies, covering different auxiliary modalities, category-disjoint zero-shot settings, and multiple RGB-based zero-shot anomaly detectors, demonstrate consistent improvements and validate the effectiveness and plug-and-play capability of the proposed framework.
\end{itemize}

\begin{figure*}[t]
\centering
\includegraphics[width=0.8\textwidth]{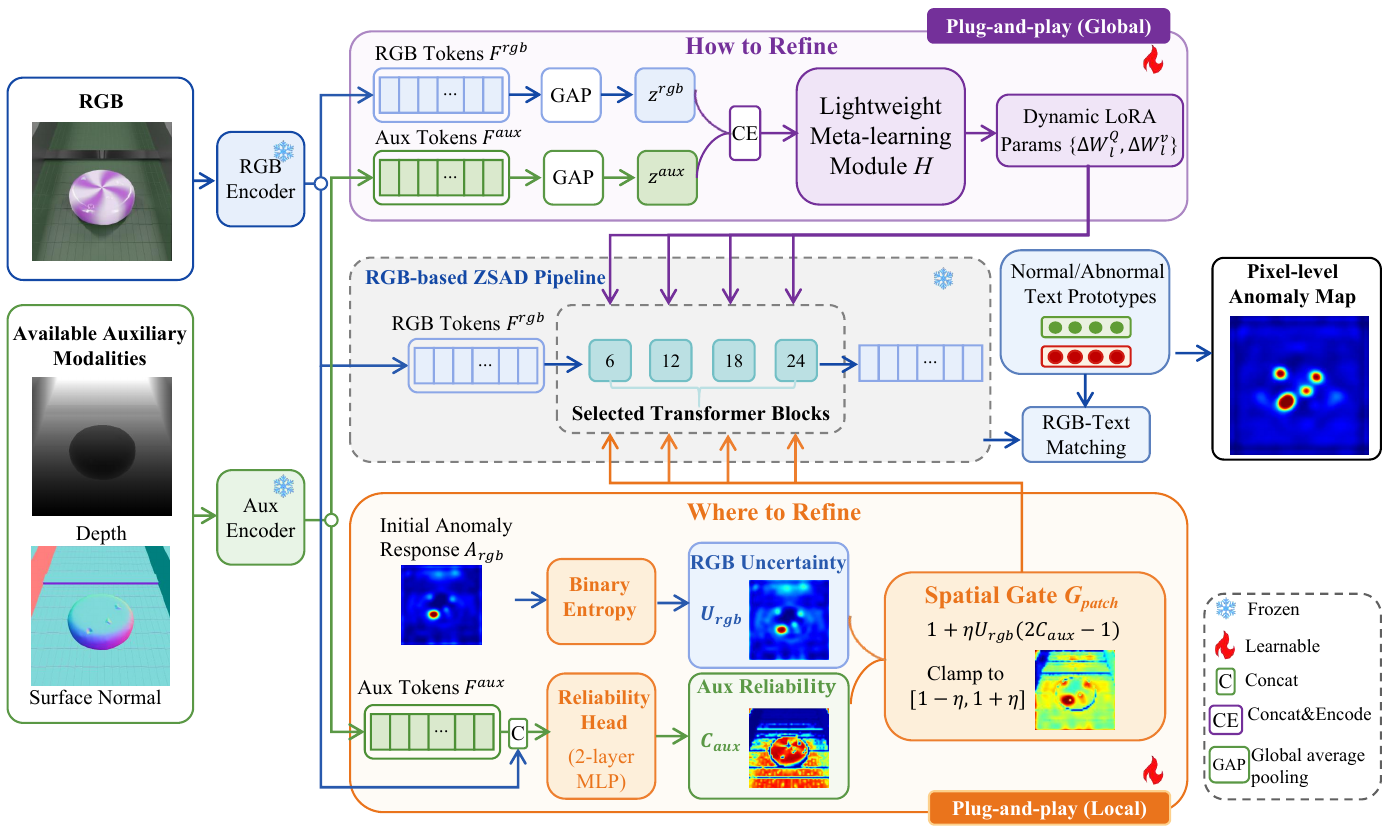}
\caption{Overall framework of our proposed method. 
}
    \vspace{-0.2cm}
\label{fig:framework}
\end{figure*}

\section{Related Work}
\subsection{Image Anomaly Detection}
Prior single-modal image anomaly detection methods mainly identify deviations from normal patterns through reconstruction, feature memory, probability density estimation, or teacher-student feature discrepancies. Representative methods, including DRAEM~\cite{zavrtanik2021draem}, PatchCore~\cite{roth2022patchcore}, CFLOW-AD~\cite{gudovskiy2022cflow}, and STFPM~\cite{wang2021student}, perform well when object categories are fixed and sufficient normal data are available. However, their anomaly decisions are typically constrained by the normal patterns observed during training. More recently, vision-language foundation models have advanced zero-shot anomaly detection. Methods such as WinCLIP~\cite{jeong2023winclip} and AnomalyCLIP~\cite{zhou2024anomalyclip} exploit pretrained image-text spaces to match local image features with normal and abnormal text prompts, enabling anomaly recognition and localization on unseen categories. Subsequent studies further improve cross-category generalization through prompt optimization, visual adaptation, or stronger visual foundation models. However, these methods still primarily rely on RGB appearance and may not fully capture anomaly cues related to geometric structure, surface properties, or material variations.

\subsection{Multimodal Image Anomaly Detection}
Multimodal image anomaly detection introduces auxiliary observations, such as depth maps, surface normals, and point clouds, to complement RGB in perceiving geometric structures and surface properties. Most existing methods exploit multimodal information under the normality modeling paradigm. EasyNet~\cite{chen2023easynet} and M3DM~\cite{wang2023multimodal} integrate multimodal features through attention or hybrid fusion strategies, while the shape-guided dual-memory learning method~\cite{chu2023shape} models normal appearance and geometry using separate memories. CFM~\cite{costanzino2024multimodal} learns cross-modal feature mappings under normal conditions, and Cycle-CFM~\cite{shi2025cycle} further improves mapping consistency through bidirectional constraints. These methods enhance defect localization by leveraging complementary modalities, but they are mainly designed for training-based anomaly detection and rely on normal samples to construct multimodal representations.

Recently, a few works have extended vision-language models to multimodal or 3D zero-shot anomaly detection. PointAD~\cite{zhou2024pointad} explores zero-shot 3D anomaly understanding by combining point clouds and images, while ZUMA~\cite{ma2026zuma} builds a joint representation of RGB and 3D information through cross-domain calibration and dynamic semantic interaction. Although these methods improve multimodal zero-shot anomaly perception, they still tend to treat RGB and auxiliary modalities as comparable semantic inputs and perform prediction based on fused or joint representations. In contrast, our method does not replace the RGB image-text anomaly prediction pathway. It serves as a plug-and-play auxiliary modulation module for RGB-based foundation zero-shot anomaly detectors, using auxiliary modalities as conditional signals to enhance RGB feature refinement without constructing a joint multimodal anomaly semantic space.

\section{Method}

\subsection{Framework Overview}
Given the $i$-th RGB image $x_i^{\mathrm{rgb}}$ and its available auxiliary modality set $\mathcal{X}_i^{\mathrm{aux}}=\{x_i^{m}\mid m\in\mathcal{A}_i\}$, where $\mathcal{A}_i$ denotes the subset of auxiliary modalities available for the current sample, the input may include depth, surface normals, or any combination of other auxiliary modalities. For clarity, the following sections describe the implementation mainly using depth and surface normals as examples, although the proposed framework does not depend on a fixed modality configuration. As shown in Figure~\ref{fig:framework}, our method is built upon an existing RGB-based foundation ZSAD framework and preserves its original RGB image-text anomaly detection pathway. Instead of constructing a unified multimodal anomaly semantic space, we treat auxiliary modalities as conditional signals that selectively refine RGB anomaly perception. Consequently, auxiliary modalities never participate directly in anomaly semantic matching, but only modulate the RGB visual representation before the final RGB image-text similarity computation.

Specifically, we formulate auxiliary enhancement as a global-to-local modulation process. At the global level, the global representations of RGB and the auxiliary modalities jointly condition a lightweight meta-learning network, which dynamically generates sample-adaptive LoRA parameters for selected Transformer layers, determining how RGB features should be refined. At the local level, the initial anomaly response of the frozen RGB foundation model is used to estimate prediction uncertainty, which is further combined with an auxiliary reliability map to construct uncertainty-aware spatial modulation. The generated LoRA residuals are strengthened at locations where RGB predictions are uncertain and auxiliary observations provide reliable support, while unreliable local interventions are suppressed. 
Finally, the refined RGB features are matched with normal and abnormal textual prototypes following the original RGB image-text matching pipeline to produce the final anomaly map. Throughout both training and inference, the RGB foundation model remains frozen, while only the lightweight conditional modulation modules are optimized, making the proposed framework a plug-and-play enhancement that can be seamlessly integrated into different RGB-based ZSAD models.

\subsection{Condition-driven Dynamic LoRA Modulation}
To introduce auxiliary information without changing the main structure of RGB-based ZSAD models, we use a lightweight meta-learning module to generate dynamic LoRA parameters. Compared with directly fine-tuning the visual encoder or using fixed LoRA, the meta-learning module generates sample-adaptive low-rank residual updates conditioned on the RGB state and the auxiliary state of the current sample. In this way, it determines the refinement direction of RGB features with only a small number of trainable parameters.

Specifically, given an RGB image $x^{\mathrm{rgb}}$ and a set of available auxiliary modalities $\mathcal{X}^{\mathrm{aux}}$, the RGB image is first fed into the RGB visual encoder of the base ZSAD model to obtain patch token features $F^{\mathrm{rgb}}$. 
Each auxiliary modality is processed by its corresponding auxiliary encoder or projection module to obtain token features aligned with the RGB tokens. When multiple auxiliary modalities are available, their features are mapped to a unified dimension $d_{\mathrm{aux}}$ and aggregated as the auxiliary token representation $F^{\mathrm{aux}}$. 

We then apply global average pooling (GAP) to the RGB tokens and auxiliary tokens, respectively, to obtain their global representations:
\begin{equation}
z^{\mathrm{rgb}}=\mathrm{GAP}(F^{\mathrm{rgb}}),\qquad
z^{\mathrm{aux}}=\mathrm{GAP}(F^{\mathrm{aux}}).
\end{equation}
The two representations are concatenated and encoded into a joint condition vector:
\begin{equation}
s=\phi([z^{\mathrm{rgb}};z^{\mathrm{aux}}]),
\end{equation}
where $[\,;\,]$ denotes channel-wise concatenation, and $\phi(\cdot)$ is a two-layer MLP (multilayer perceptron). The condition vector $s$ is further sent to a lightweight meta-learning module $H(\cdot)$ to dynamically generate LoRA parameters for selected Transformer layers:
\begin{equation}
\{\Delta W_l^Q,\Delta W_l^V\}_{l\in\mathcal{L}}=H(s),
\end{equation}
where $H(\cdot)$ is a three-layer MLP that maps the joint condition $s$ to the LoRA parameters of all target layers, and $\mathcal{L}$ denotes the set of target layers equipped with dynamic LoRA. For the input tokens $X_l$ of the $l$-th layer, the query and value residuals generated by dynamic LoRA are defined as:
\begin{equation}
\Delta Q_l=X_l\Delta W_l^Q,\qquad
\Delta V_l=X_l\Delta W_l^V.
\end{equation}
The global RGB–auxiliary condition therefore determines the sample-specific refinement direction in the low-rank parameter space. Importantly, these residuals do not directly replace the original RGB features and are further controlled by the spatial modulation mechanism introduced in the next subsection. This design allows auxiliary observations to guide RGB anomaly perception through parameter-efficient modulation, without directly participating in anomaly-semantic construction or RGB image-text similarity computation.

\subsection{Uncertainty-aware Spatial Modulation}

The dynamic LoRA residuals generated by global conditions provide a sample-adaptive refinement direction, but auxiliary information is not equally useful at all local regions. We therefore introduce an uncertainty-aware spatial modulation mechanism to control the strength of LoRA residual updates at the patch level. The key idea is to strengthen residual updates where RGB anomaly prediction is uncertain and auxiliary reliability is high, while suppressing unreliable local intervention.

We first estimate the initial anomaly response from the RGB image-text matching result of the RGB-based ZSAD model. Given the RGB patch feature $f_j^{\mathrm{rgb}}$ and the normal and abnormal text prototypes $T$, the probability that the $j$-th patch belongs to the abnormal class is defined as:
\begin{equation}
A_{\mathrm{rgb}}(j)=p(y=\mathrm{abnormal}\mid f_j^{\mathrm{rgb}},T).
\end{equation}
Let $a_j=A_{\mathrm{rgb}}(j)$. We measure the uncertainty of the RGB prediction using normalized binary entropy:
\begin{equation}
\begin{array}{l}
U_{\mathrm{rgb}}(j)
=
-\frac{1}{\log 2}
\Big[
a_j\log(a_j+\epsilon) \\[2pt]
\quad +(1-a_j)\log(1-a_j+\epsilon)
\Big],
\end{array}
\end{equation}
where $\epsilon$ is a small constant used for numerical stability. A larger $U_{\mathrm{rgb}}(j)$
indicates greater ambiguity between the normal and abnormal classes, so the patch may benefit more
from auxiliary modulation.

In parallel, we estimate the patch-wise reliability of the auxiliary modalities using a lightweight reliability prediction head. The RGB tokens and aggregated auxiliary tokens are concatenated along the channel dimension and fed into $\phi_c$:
\begin{equation}
C_{\mathrm{aux}}
=
\sigma
\left(
\phi_{c}
\left(
[F^{\mathrm{rgb}};F^{\mathrm{aux}}]
\right)
\right),
\end{equation}
where $\phi_{c}$ is a two-layer MLP, and $\sigma(\cdot)$ is the sigmoid function. 

\begin{table*}[t]
\centering
{\small
\setlength{\tabcolsep}{4.2pt}
\begin{tabular}{@{}lcccccccc@{}}
\toprule
& \multicolumn{4}{c}{MVTec 3D-AD}
& \multicolumn{4}{c}{Eyecandies} \\
\cmidrule(lr){2-5}
\cmidrule(lr){6-9}
Method
& P-AUC (\%) & AUPRO (\%) & I-AUC (\%) & AP (\%)
& P-AUC (\%) & AUPRO (\%) & I-AUC (\%) & AP (\%) \\
\midrule

CLIP + R.
& -- & 56.0 & 60.4 & 86.4
& 78.0 & 31.8 & 73.0 & 73.9 \\

PointCLIP V2
& 78.3 & 49.4 & 49.8 & 79.3
& 46.0 & -- & 46.9 & 49.9 \\

PointCLIP V2a
& 79.5 & 51.6 & 49.4 & 79.8
& 46.2 & -- & 48.5 & 50.5 \\

PointAD-CoOp
& 96.5 & 88.8 & 83.4 & 94.9
& 94.9 & 83.6 & 73.7 & 76.0 \\

PointAD
& 97.2 & 90.2 & 86.9 & \underline{96.1}
& 95.3 & 84.3 & 77.7 & 80.4 \\

ZUMA
& 97.6 & 92.5 & 84.0 & --
& 97.3 & 88.9 & 74.9 & -- \\

ZUMA-FT
& \underline{97.8} & \underline{93.1}
& \underline{88.2} & --
& \underline{97.5} & 89.4
& 78.2 & -- \\

\midrule

AnomalyCLIP + Ours
& 95.6 & 84.3 & 76.5 & 92.2
& 95.4 & 81.5
& \underline{81.0} & \underline{81.1} \\

AA-CLIP + Ours
& 97.4 & 90.9 & 76.5 & 92.6
& 97.4 & \underline{89.8}
& 78.3 & 80.7 \\

AnomalyVFM + Ours
& \textbf{98.5} & \textbf{94.4}
& \textbf{88.5} & \textbf{96.5}
& \textbf{98.2} & \textbf{91.7}
& \textbf{88.0} & \textbf{89.2} \\

\bottomrule
\end{tabular}
}
\caption{Comparison with state-of-the-art methods on MVTec 3D-AD and Eyecandies.
The best and second-best results within each dataset and metric are highlighted in bold and underlined, respectively. ``--'' denotes unavailable results.}
\label{tab:main_results}
\end{table*}

Based on RGB uncertainty and auxiliary reliability, we construct a spatial modulation map centered at one:
\begin{equation}
G_{\mathrm{patch}}
=
\mathrm{clip}
\left(
1+\eta\cdot U_{\mathrm{rgb}}\cdot(2C_{\mathrm{aux}}-1),
1-\eta,
1+\eta
\right),
\end{equation}
where $\eta\in[0,1)$ controls the modulation range, and $\mathrm{clip}(\cdot)$ restricts the modulation value within $[1-\eta,1+\eta]$. This residual-style formulation constrains the modulation around the original LoRA response, allowing auxiliary information to enhance or suppress local residual updates without disrupting the original adaptation behavior. When the RGB prediction is confident, $U_{\mathrm{rgb}}\approx0$ and thus $G_{\mathrm{patch}}\approx1$. When RGB is uncertain and the auxiliary reliability is high, $C_{\mathrm{aux}}>0.5$ and $G_{\mathrm{patch}}>1$, which strengthens the local residual update. Conversely, when RGB is uncertain but auxiliary reliability is low, $C_{\mathrm{aux}}<0.5$ and $G_{\mathrm{patch}}<1$, which suppresses unreliable auxiliary intervention.

Because the spatial modulation is designed for local patch representations, we exclude the class token from patch-wise reliability modulation. Specifically, a fixed factor of one is assigned to the class token:
\begin{equation}
G=[\mathbf{1}_{\mathrm{cls}};G_{\mathrm{patch}}],
\end{equation}
thereby preserving the global image representation while selectively modulating local patch tokens. For the target Transformer layer $l$, the resized modulation factor $G_l$ is applied to the dynamic LoRA residuals:
\begin{equation}
\Delta Q_l'=G_l\odot\Delta Q_l,
\qquad
\Delta V_l'=G_l\odot\Delta V_l.
\end{equation}
The modulated residuals are then added to the original query and value projections, allowing dynamic LoRA modulation to determine how to refine RGB features and uncertainty-aware spatial modulation to determine where to strengthen or suppress the residuals.

\subsection{Plug-and-play Optimization and Inference}
Our framework retains the original optimization objective of each RGB-based ZSAD model and introduces no additional multimodal loss. The RGB visual encoder and other task-specific modules remain frozen, while only the proposed conditional modulation modules are optimized through the base detector objective. During inference, global RGB and auxiliary representations generate sample-adaptive LoRA parameters, whose residual updates are spatially modulated and injected into selected Transformer layers. The refined RGB features are then matched with the original normal and abnormal text prototypes to produce image-level and pixel-level anomaly predictions. Since the RGB image-text matching pathway remains unchanged, the proposed method can be directly integrated into different RGB-based zero-shot anomaly detectors.

\begin{table*}[t]
\centering
{\small
\setlength{\tabcolsep}{3.2pt}
\begin{tabular}{@{}llcccccccc@{}}
\toprule
\multirow{2}{*}{Framework}
& \multirow{2}{*}{Config.}
& \multicolumn{4}{c}{MVTec 3D-AD}
& \multicolumn{4}{c}{Eyecandies} \\
\cmidrule(lr){3-6}
\cmidrule(lr){7-10}
&
& P-AUC (\%) & AUPRO (\%) & I-AUC (\%) & AP (\%)
& P-AUC (\%) & AUPRO (\%) & I-AUC (\%) & AP (\%) \\
\midrule

\multirow{4}{*}{AnomalyCLIP}

& RGB baseline
& 95.2 & 82.0 & 71.8 & 90.2
& 91.1 & 67.7 & 65.1 & 65.9 \\

& RGB + Direct Fusion
& 93.0 & 74.0 & 72.2 & 90.3
& 91.3 & 73.1 & 73.6 & 74.8 \\

& + Dyn. LoRA
& 95.5 & 82.9 & 75.4 & 91.8
& 95.0 & 79.5 & 80.3 & 80.7 \\

& + Dyn. LoRA + S. Mod.
& \textbf{95.6} & \textbf{84.3}
& \textbf{76.5} & \textbf{92.2}
& \textbf{95.4} & \textbf{81.5}
& \textbf{81.0} & \textbf{81.1} \\

\midrule

\multirow{4}{*}{AA-CLIP}
& RGB baseline
& 96.6 & 90.9 & 74.7 & 92.0
& 96.7 & 86.8 & 72.7 & 74.7 \\

& RGB + Direct Fusion
& 95.6 & 88.3 & 75.1 & 92.0
& 95.1 & 84.6 & 73.2 & 75.2 \\

& + Dyn. LoRA
& 97.4 & \textbf{91.5}
& 75.8 & 92.3
& 95.7 & 84.5 & 74.2 & 75.7 \\

& + Dyn. LoRA + S. Mod.
& \textbf{97.4} & 90.9
& \textbf{76.5} & \textbf{92.6}
& \textbf{97.4} & \textbf{89.8}
& \textbf{78.3} & \textbf{80.7} \\

\midrule

\multirow{4}{*}{AnomalyVFM}
& RGB baseline
& 97.8 & 91.9
& 88.5 & 96.2
& 90.2 & 79.9 & 80.4 & 82.8 \\

& RGB + Direct Fusion
& 98.0 & 92.5 & 87.5 & 96.1
& 93.4 & 83.4 & 80.7 & 82.8 \\

& + Dyn. LoRA
& 98.4 & 94.3 & 87.7 & 96.1
& 98.0 & 91.5 & 87.7 & 89.2 \\

& + Dyn. LoRA + S. Mod.
& \textbf{98.5} & \textbf{94.4}
& \textbf{88.5} & \textbf{96.5}
& \textbf{98.2} & \textbf{91.7}
& \textbf{88.0} & \textbf{89.2} \\

\bottomrule
\end{tabular}
}
\caption{Component-wise ablation results on MVTec 3D-AD and Eyecandies. The best result for each framework, dataset, and metric is highlighted in bold.}
\label{tab:module_ablation}
\end{table*}

\begin{table*}[t]
\centering
{\small
\setlength{\tabcolsep}{4.2pt}
\begin{tabular}{@{}lccccccccc@{}}
\toprule
\multirow{2}{*}{Framework}
& \multirow{2}{*}{$\eta$}
& \multicolumn{4}{c}{MVTec 3D-AD}
& \multicolumn{4}{c}{Eyecandies} \\
\cmidrule(lr){3-6}
\cmidrule(lr){7-10}
&
& P-AUC (\%) & AUPRO (\%) & I-AUC (\%) & AP (\%)
& P-AUC (\%) & AUPRO (\%) & I-AUC (\%) & AP (\%) \\
\midrule

\multirow{5}{*}{AnomalyCLIP}
& 0
& 95.5 & 82.9 & 75.4 & 91.8
& 95.0 & 79.5 & 80.3 & 80.7 \\

& 0.05
& \textbf{95.6} & 83.6 & 76.4 & 92.1
& 94.6 & 79.4 & 79.3 & 80.3 \\

& 0.10
& \textbf{95.6} & 83.6 & 76.4 & \textbf{92.2}
& \textbf{95.4} & 81.2 & 80.6 & \textbf{81.1} \\

& 0.15
& \textbf{95.6} & 84.3 & \textbf{76.5} & \textbf{92.2}
& \textbf{95.4} & \textbf{81.5} & \textbf{81.0} & \textbf{81.1} \\

& 0.20
& 95.3 & \textbf{84.8} & 75.6 & 91.6
& 94.8 & 79.6 & 80.6 & 80.7 \\

\midrule

\multirow{5}{*}{AnomalyVFM}
& 0
& 98.4 & 94.3 & 87.7 & 96.1
& 98.0 & 91.5 & 87.7 & 89.2 \\

& 0.05
& \textbf{98.5} & \textbf{94.4} & 87.9 & 96.2
& \textbf{98.2} & 91.7 & \textbf{88.3} & \textbf{89.6} \\

& 0.10
& \textbf{98.5} & \textbf{94.4} & 87.9 & 96.4
& 98.1 & 91.7 & 87.8 & 89.2 \\

& 0.15
& \textbf{98.5} & \textbf{94.4} & \textbf{88.5} & \textbf{96.5}
& \textbf{98.2} & 91.7 & 88.0 & 89.2 \\

& 0.20
& 98.4 & 94.3 & 87.3 & 96.0
& \textbf{98.2} & \textbf{92.0} & 87.8 & 89.0 \\

\bottomrule
\end{tabular}
}
\caption{Sensitivity analysis of the spatial modulation amplitude $\eta$ on MVTec 3D-AD and Eyecandies.}
\vspace{-0.3cm}
\label{tab:eta_ablation}
\end{table*}

\section{Experiments}
\subsection{Datasets and Evaluation Metrics}
\subsubsection{Datasets.}
MVTec 3D-AD~\cite{bergmann2021mvtec} targets real industrial inspection and contains 10 object categories. Each sample provides an RGB image and a corresponding 3D observation, covering anomalies such as scratches, dents, contamination, holes, and structural deformation. 
Eyecandies~\cite{bonfiglioli2022eyecandies} is a synthetic multimodal benchmark with 10 candy object categories. It provides RGB, depth, and surface normal modalities under complex textures, self-occlusions, and specular reflections. 

\subsubsection{Evaluation metrics.}
We report four metrics: Pixel-level AUROC (P-AUC), Area Under the Per-Region Overlap (AUPRO), Image-level AUROC (I-AUC), and image-level Average Precision (AP). P-AUC and AUPRO evaluate pixel-level anomaly localization, while I-AUC and AP measure image-level anomaly detection.

\subsection{Experimental Details}
We integrate our plug-and-play method into three representative anomaly detection frameworks, including AnomalyCLIP, AA-CLIP, and AnomalyVFM. For each baseline, we initialize the model with its VisA-pretrained checkpoint and freeze all parameters of the original framework during training, including the visual encoder, text encoder, existing adaptation modules, and prediction head or decoder. Following prior works~\cite{zhou2024pointad,ma2026zuma}, we adopt the one-vs-rest protocol for category-disjoint zero-shot evaluation and report results averaged over three target-category splits. Dynamic LoRA modules are inserted into selected attention layers of the visual Transformer and applied to the Query and Value projections. In all experiments, the LoRA rank is set to $4$, the auxiliary dimension $d_{\mathrm{aux}}$ is set to 64, and the learning rate of newly introduced modules is set to $1\times10^{-4}$. Unless otherwise specified, we use the full model with spatial modulation amplitude $\eta=0.15$. All experiments are conducted on a single NVIDIA RTX 3090 GPU.

\subsection{Main Results}
Table~\ref{tab:main_results} compares our method with existing multimodal ZSAD methods on MVTec 3D-AD and Eyecandies, including CLIP + Rendering (CLIP + R.)~\citep{radford2021learning}, PointCLIP V2/V2a~\citep{zhang2022pointclip}, PointAD-CoOp/PointAD~\citep{zhou2022learning,zhou2024pointad}, and ZUMA/ZUMA-FT~\citep{ma2026zuma}. We also integrate our modules into AnomalyCLIP, AA-CLIP, and AnomalyVFM to evaluate their generality across different frameworks. 
On both benchmarks, AnomalyVFM + Ours achieves the best performance across all metrics. Specifically, on Eyecandies, it outperforms the previous best methods by 0.7\%, 2.3\%, 9.8\%, and 8.8\% in P-AUC, AUPRO, I-AUC, and AP, respectively. 
When integrated with other detectors, our framework consistently improves AnomalyCLIP and AA-CLIP, achieving second-best performance on several metrics. These results demonstrate that the proposed auxiliary-conditioned modulation is not tied to a specific detector and can provide consistent cross-model benefits.

\begin{figure*}[!t]
    \centering
    \includegraphics[width=0.8\textwidth]{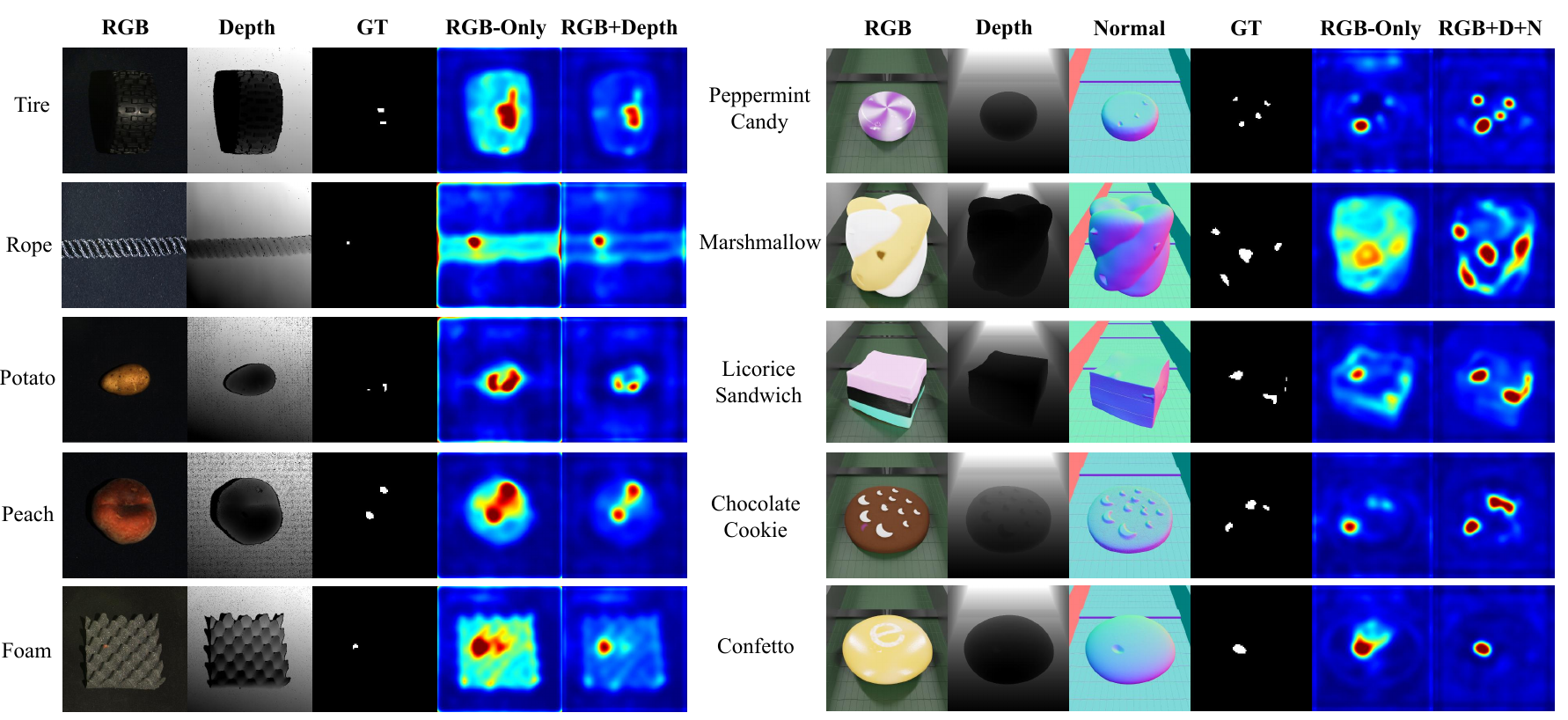}
    \caption{Qualitative anomaly localization results on five representative categories from MVTec 3D-AD (left) and Eyecandies (right), using AnomalyVFM as the base detector. Each sample shows the input modalities, ground-truth mask, RGB-only anomaly map, and auxiliary-conditioned final anomaly map.}
    \label{fig:vis_results}
        \vspace{-0.2cm}
\end{figure*}

\subsection{Ablation Study}
\subsubsection{Module Ablation.}
We ablate dynamic LoRA modulation and uncertainty-aware spatial modulation on frozen RGB baselines, and further compare them with a direct feature fusion strategy under the same backbone and protocol. In Table~\ref{tab:module_ablation}, Dyn. LoRA and S. Mod. denote these two modules, respectively. All variants within the same framework use the same pretrained parameters, and only newly introduced modules are optimized. We use depth on MVTec 3D-AD and depth+surface normal on Eyecandies. As shown in Table~\ref{tab:module_ablation}, direct fusion provides limited or inconsistent improvements, whereas the proposed modulation strategy consistently enhances anomaly localization and image-level detection. For instance, the full AnomalyCLIP variant improves AUPRO and I-AUC by $13.8\%$ and $15.9\%$ over the RGB baseline on Eyecandies. For AA-CLIP, Dynamic LoRA alone slightly decreases P-AUC and AUPRO, while adding Spatial Mod. improves AUPRO and I-AUC by $3.0\%$ and $5.6\%$ over the baseline. For AnomalyVFM on Eyecandies, Dynamic LoRA modulation improves P-AUC and AUPRO by $7.8\%$ and $11.6\%$. These results demonstrate that Dynamic LoRA incorporates auxiliary geometric cues, while uncertainty-aware spatial modulation controls their local influence and reduces negative cross-modal transfer.

\subsubsection{Sensitivity to the Spatial Modulation Amplitude.}
The spatial modulation amplitude $\eta$ controls the strength of uncertainty-aware spatial modulation, where $\eta=0$ denotes Dynamic LoRA without the uncertainty-aware spatial modulation. We vary $\eta$ from 0 to 0.2 while keeping all other settings unchanged. As shown in Table~\ref{tab:eta_ablation}, the uncertainty-aware spatial modulation generally brings additional gains over Dynamic LoRA alone, although the optimal value varies across metrics. Overall, $\eta=0.15$ achieves the most balanced performance across datasets and frameworks, while larger values tend to favor pixel-level localization and smaller values may yield better image-level discrimination. Therefore, we set $\eta=0.15$ as the default setting in all experiments without dataset-specific tuning.

\subsection{Visualization}
Figure~\ref{fig:vis_results} presents qualitative comparisons on MVTec 3D-AD and Eyecandies. Compared with the RGB-only baseline AnomalyVFM, AnomalyVFM + Ours produces more concentrated anomaly responses within ground-truth defect regions and shows better consistency with the corresponding masks. On MVTec 3D-AD, depth cues help suppress background responses and refine localization for geometric defects. On Eyecandies, depth and surface normal cues further highlight structural and surface anomalies that are difficult to distinguish from RGB appearance alone. These qualitative results show that auxiliary-conditioned modulation improves pixel-level anomaly localization while retaining the original RGB image-text prediction pathway.

\begin{figure}[t]
    \centering
    \includegraphics[width=\columnwidth]{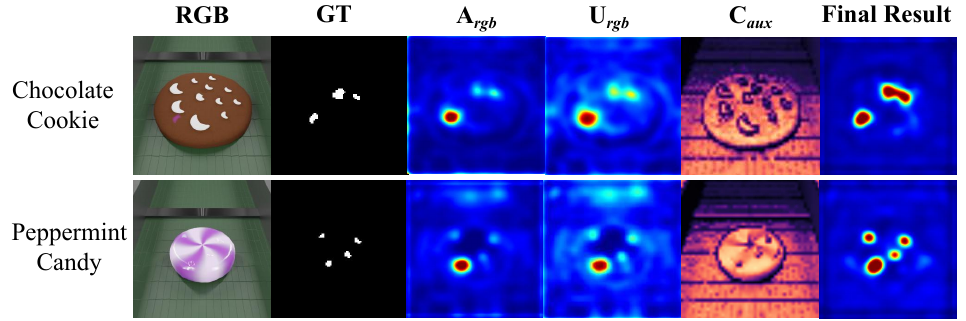}
    \caption{Visualization of uncertainty-aware spatial modulation on two Eyecandies samples.} 
    \label{fig:vis_map}
\end{figure}

To further interpret the proposed uncertainty-aware spatial modulation, Figure~\ref{fig:vis_map} visualizes the intermediate maps on two Eyecandies samples. The initial RGB anomaly response $A_{\mathrm{rgb}}$ provides a coarse anomaly estimate, while the uncertainty map $U_{\mathrm{rgb}}$ highlights ambiguous regions where RGB prediction is less confident. The auxiliary reliability map $C_{\mathrm{aux}}$ identifies regions where auxiliary modalities provide reliable structural cues. By combining these signals, the spatial modulation selectively refines uncertain but auxiliary-supported regions, leading to the final anomaly results that better align with the ground-truth masks.

\section{Conclusion}
We present a plug-and-play auxiliary-conditioned framework for zero-shot anomaly detection. Instead of constructing a joint multimodal anomaly space, our method preserves the original RGB image-text matching pathway and uses auxiliary modalities to selectively refine RGB anomaly perception. A lightweight meta-learning module generates sample-adaptive LoRA residuals from global RGB and auxiliary representations, while uncertainty-aware spatial modulation controls their local activation using the initial RGB anomaly response and auxiliary reliability. This global-to-local design introduces complementary structural and surface cues without disrupting pretrained RGB anomaly semantics. Extensive experiments on MVTec 3D-AD and Eyecandies demonstrate consistent improvements across different auxiliary modalities and RGB-based detectors, confirming the effectiveness and generality of the proposed framework.

\bibliography{aaai2027}


\end{document}